\documentclass{article}
\usepackage{spconf,amsmath,graphicx,hyperref, color}

\title{Low-Latency LLM-Based Streaming Multi-Talker ASR with Verifiable Rewards}
\name{Jing Pan \qquad Ruizhi Li \qquad Ju Lin \qquad Jing Zheng \qquad Florian Metze}

\address{Meta, USA}
\begin{document}
%
\maketitle
\begin{abstract}
LLM-based architectures offer strong generalization and instruction-following capabilities for streaming automatic speech recognition (ASR), but multi-talker use introduces coupled transcription, attribution, and latency challenges. Our wearable setting requires real-time transcription of a conversation between a near-field device wearer and a far-field partner. Because partial hypotheses are rendered on persistent sides of a dialogue UI, speaker roles cannot be permuted and each speaker tag must precede its turn. These sparse tags are weakly optimized by supervised fine-tuning (SFT), while committing to a leading tag can delay the text that follows. We therefore post-train a streaming LLM-based ASR model with reinforcement learning with verifiable rewards (RLVR), combining a fixed-role Speaker-Attributed WER (SA-WER) grader with a turn-level onset-delay reward. On an eight-direction multi-talker cohort, SA-WER RL reduces SA-WER from $20.51\%$ for SFT to $13.14\%$. Adding the delay objective reduces onset $p50$ by $65\%$ (from $0.918\,\text{s}$ to $0.320\,\text{s}$) and $p90$ by $44\%$ (from $1.389\,\text{s}$ to $0.780\,\text{s}$) relative to SA-WER-only RL, while achieving $12.70\%$ SA-WER. These results demonstrate that fixed-role recognition quality and streaming responsiveness can be improved jointly.
\end{abstract}
\begin{keywords}
streaming speech recognition, multi-talker, reinforcement learning, verifiable reward
\end{keywords}
\section{Introduction}
\begin{figure}[t]
    \centering
    \includegraphics[width=\columnwidth]{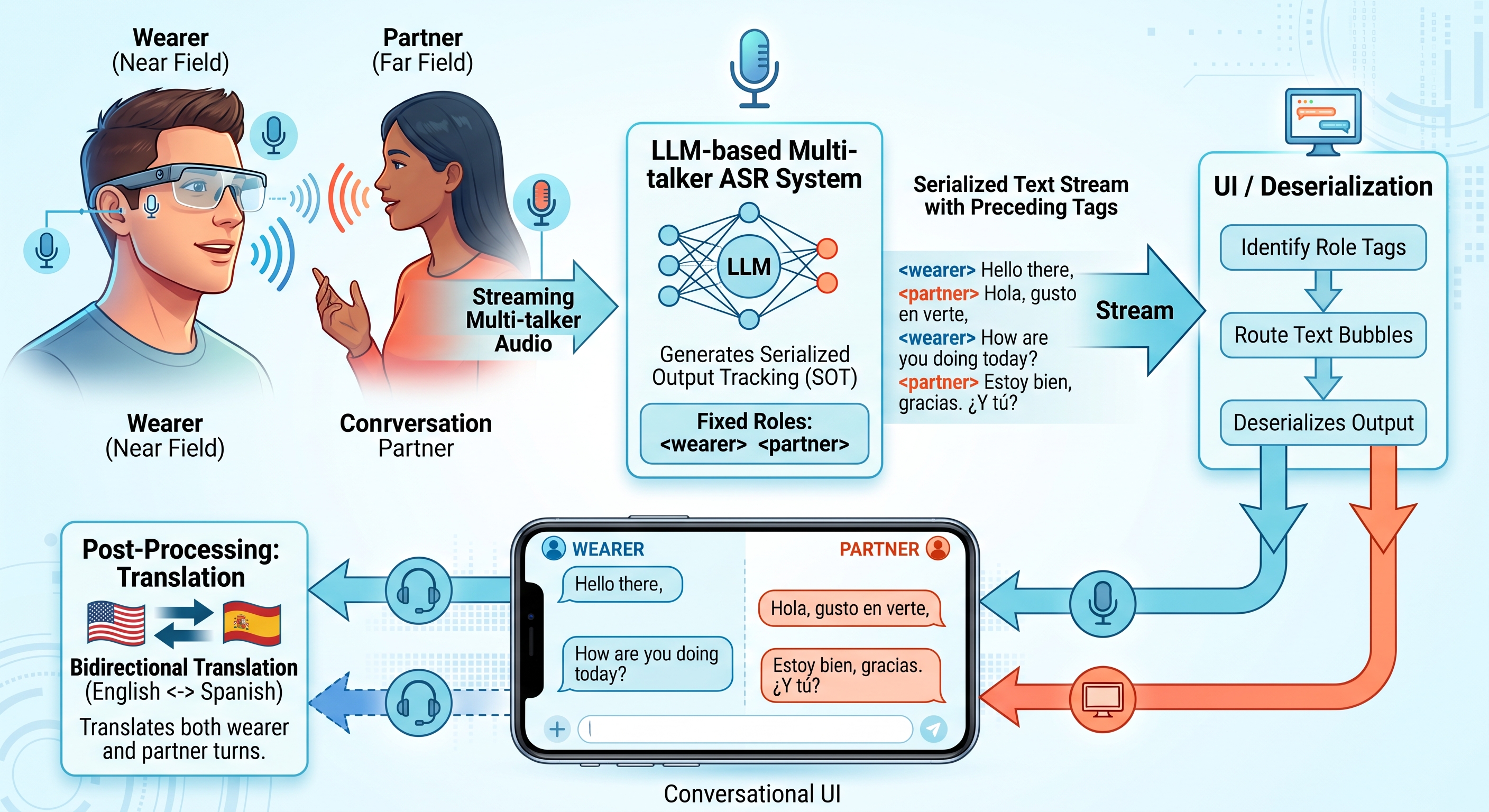}
    \caption{Wearable two-party streaming pipeline with fixed-role SOT, dialogue-side routing, and downstream bidirectional translation.}
    \label{fig:wearable-multitalker-pipeline}
\end{figure}

Large language model (LLM)-based architectures are increasingly used for automatic speech recognition (ASR), motivated by their generalization, contextual modeling, and instruction-following capabilities \cite{Wu2023,Sun2024,Yang2024, Pan2024, abouelenin2025phi}. Recent streaming speech LLMs have demonstrated competitive transcription and translation quality under real-time constraints \cite{shi2026,Ma2026,liu2026voxtral}.
Extending this paradigm to multi-talker ASR requires both recognizing overlapping or alternating speech and determining \emph{who spoke what}. Prior streaming systems address these challenges with dedicated multi-channel or speech-unmixing architectures, including specialized Recurrent Neural Network Transducers (RNN-Ts) \cite{Lu2021}. Meng et al. \cite{meng2025large} demonstrates that speech LLM can understand overlapped human speech following versatile human instructions. Serialized Output Training (SOT) instead linearizes multiple speakers' transcriptions into one sequence separated by speaker-change tokens \cite{Kanda2020}, making it naturally compatible with autoregressive LLM decoders. Our wearable use case transcribes a real-time two-party conversation between the device wearer (near field) and a conversation partner (far field), while the UI deserializes the output stream into the two sides of a dialogue, as illustrated in Fig.~\ref{fig:wearable-multitalker-pipeline}. Speaker identities must therefore be fixed semantic roles---\texttt{<wearer>} and \texttt{<partner>}---rather than labels that may be permuted during scoring. Moreover, each role tag must \emph{precede} its turn so the UI can route partial words to the correct side as they arrive; a suffix tag would require buffering the complete turn before displaying it. Recent LLM-based multi-talker ASR combines neural beamforming with SOT supervision \cite{lin2026equipping}, but is not natively streaming and requires an additional streaming harness.

Supervised Fine-Tuning (SFT) can teach this SOT format, but token-level cross-entropy is inefficient for optimizing multi-talker recognition quality. It does not directly optimize sequence-level metrics such as word error rate (WER), and sparse speaker tags contribute little to the total loss. A model can therefore achieve low lexical loss while omitting the quieter speaker or assigning words to the wrong role. Reinforcement learning with verifiable rewards (RLVR) is a natural post-training alternative because ASR hypotheses admit automatic, quantitative scoring against references. Prior work has explored sequence-level reinforcement learning for ASR and reward-based post-training for speech models \cite{Tjandra2019,Gao2025}. We use group-relative reward normalization, following Group Relative Policy Optimization (GRPO) \cite{Shao2024}, to optimize multiple sampled transcripts per audio prompt.

These constraints motivate two rewards. Because cpWER \cite{vonNeumann2025} can permute speakers to minimize error, we use fixed-role Speaker-Attributed WER (SA-WER) to score transcription and attribution. Since each turn must begin with a role tag, the model may delay all text while confirming the speaker; an onset-delay reward encourages earlier commitment without sacrificing recognition quality. Our contributions are: 1) \textbf{Fixed-role quality alignment:} We formulate an RLVR objective around a fixed-role SA-WER grader, directly optimizing transcription and speaker attribution for streaming SOT output. 2) \textbf{Coverage-aware latency optimization:} We introduce a bounded onset delay factor that rewards timely emission of aligned turns without allowing latency to dominate quality.

On our 2-speaker conversational evaluation sets, RLVR with SA-WER grader reduced the streaming SA-WER from $20.51\%$ for SFT to $13.14\%$. Adding the delay objective reduces onset latency relative to SA-WER-only RL: $p50$ decreases by $65\%$, from $0.918\,\text{s}$ to $0.320\,\text{s}$, and $p90$ by $44\%$, from $1.389\,\text{s}$ to $0.780\,\text{s}$. The resulting model achieves $12.70\%$ WER, demonstrating joint improvements in speaker-aware quality and streaming responsiveness.

\section{Methodology}
\begingroup
\setlength{\abovedisplayskip}{4pt plus 1pt minus 1pt}
\setlength{\belowdisplayskip}{4pt plus 1pt minus 1pt}
\setlength{\abovedisplayshortskip}{2pt plus 1pt}
\setlength{\belowdisplayshortskip}{3pt plus 1pt minus 1pt}
\makeatletter
\renewcommand\subsection{\@startsection{subsection}{2}{\z@}%
  {-1.8ex plus -.4ex minus -.2ex}%
  {0.6ex plus .1ex}%
  {\normalfont\normalsize\bfseries}}
\makeatother

\subsection{RLVR Framework}
We initialize the policy $\pi_\theta$ from a supervised fine-tuned model and post-train it with reinforcement learning with verifiable rewards (RLVR). For each audio prompt, we sample multiple transcript rollouts and score them with the graders described below. Our objective is similar to GRPO \cite{Shao2024}, except that it directly weights each sampled transcript's log-likelihood by its group-normalized advantage instead of using a clipped surrogate. We additionally apply full-distribution reverse KL regularization to the frozen SFT policy.

\subsection{Fixed-Role Speaker-Attributed WER}
Our SOT transcript begins each turn with a fixed-role tag for the wearer ($s=0$) or partner ($s=1$). After normalization, each word is represented as $(w,s)$. A joint edit-distance alignment compares the serialized hypothesis against separate wearer and partner reference streams without permitting role permutation. Besides insertion ($I$), deletion ($D$), and substitution ($S$), it identifies $A$, a correct word assigned to the wrong speaker, and $X$, an error in both word and speaker. For $N$ reference words,
\begin{equation}
\mathrm{SA\text{-}WER}=\frac{I+D+S+A+X}{N},\qquad
\mathrm{SAE}=\frac{A+X}{N}.
\label{eq:sawer_metrics}
\end{equation}
We also define the deletion rate $\mathrm{DR}=(D_0+D_1)/N$ to capture the overall deletion errors.

\subsection{SA-WER Grader}
The fixed-role alignment already includes attribution errors and deletion errors in $\mathrm{SA\text{-}WER}$. To emphasize these product-critical failures during RL, our grader up-weights them through the combined error
\begin{equation}
E = \frac{\mathrm{SA\text{-}WER}
+ \alpha_{\mathrm{SAE}}\,\mathrm{SAE}
+ \alpha_{\mathrm{DR}}\,\mathrm{DR}}
{1+\alpha_{\mathrm{SAE}}+\alpha_{\mathrm{DR}}},
\label{eq:combined_error}
\end{equation}
The denominator keeps the combined quantity on the scale of an error rate. We convert it to a bounded quality reward using
\begin{equation}
R_{\mathrm{SA}} = \left(1-\operatorname{clip}(E,0,1)\right)^2.
\label{eq:sawer_reward}
\end{equation}
A perfect transcript receives reward 1, while hypotheses with combined error at or above 1 receive reward 0. This sequence-level reward directly exposes lexical accuracy, fixed-role attribution, and transcription retention to RL optimization.

\subsection{Turn-Level Onset-Delay Regularization}
Our streaming decoder interleaves audio consumption and text generation. Each advancing audio marker, including empty-audio markers after stream exhaustion, increments the model-time clock by $80\,\text{ms}$. Predicted and reference turns are paired by aligning their fixed-role speaker-tag sequences and retaining only matching subsequences. Unmatched predicted or reference turns receive no latency credit and do not shift every subsequent pairing.

For a paired predicted turn $i$ and reference turn $j$, let $\hat{t}^{\mathrm{start}}_i$ be the model-time at which the leading speaker tag is emitted and $t^{\mathrm{start}}_j$ the oracle turn start. We define onset delay as
\begin{equation}
d_{ij}=\hat{t}^{\mathrm{start}}_i-t^{\mathrm{start}}_j.
\label{eq:onset_delay}
\end{equation}
For an onset buffer $b$ and decay scale $\tau$, the turn receives score
\begin{equation}
s(d_{ij};b,\tau)=\exp\left(-\frac{\max(0,d_{ij}-b)}{\tau}\right).
\label{eq:delay_score}
\end{equation}
Emission within the buffer receives full credit, while excess delay decays exponentially. Negative delay is not penalized by Eq.~\eqref{eq:delay_score}, but is tracked separately as premature emission. Let $\mathcal{M}$ be the set of matched turn pairs. The coverage-aware onset factor is
\begin{equation}
F_{\mathrm{on}} = \frac{\sum_{(i,j)\in\mathcal{M}} s(d_{ij};b,\tau)}
{\max(N_{\mathrm{hyp}},N_{\mathrm{ref}})}.
\label{eq:onset_factor}
\end{equation}
The denominator ensures that missing and hallucinated turns reduce the factor.

Finally, the onset factor acts as a bounded multiplicative regularizer on the quality reward:
\begin{equation}
\begin{aligned}
M_{\mathrm{delay}} &= \frac{1+\lambda F_{\mathrm{on}}}{1+\lambda}, \\
R_{\mathrm{final}} &= R_{\mathrm{SA}} \times M_{\mathrm{delay}}.
\end{aligned}
\label{eq:final_reward}
\end{equation}
Here, $\lambda$ controls the latency-reward strength. The multiplier equals 1 at full onset credit and is bounded below by $1/(1+\lambda)$, making onset latency a secondary regularizer on the speaker-aware recognition objective.

\endgroup

\section{Experiments}
\subsection{Experimental Setup}

\subsubsection{Base Model}
We use a streaming mixture-of-experts (MoE) speech LLM initialized from a checkpoint pretrained for multilingual single-talker ASR. Its 32 transformer layers are all MoE layers, each containing 256 routed experts with top-8 routing. The model activates approximately 5.5B backbone parameters per token. The model consumes a 24-kHz waveform and represents each $1{,}920$-sample ($80\,\text{ms}$) audio block with one audio token. We continue SFT on multi-talker conversations while retaining the fixed-role SOT tags, and then apply RLVR. This setup isolates the effects of sequence-level quality and latency optimization from changes to the model architecture or acoustic front end.

\vspace{-0.6em}
\subsubsection{Data}
\vspace{-0.2em}
The multi-talker training corpus is generated through simulation. We sample two single-talker utterances from a multilingual ASR pool and randomly assign their speakers to the wearer (near-field) and partner (far-field) roles. To reproduce the role-dependent acoustics of the wearable microphone array, we convolve the utterances with corresponding near- and far-field room impulse responses. We then apply stochastic temporal offsets and overlap the two signals to form a simulated conversation. The simulator produces five-channel audio; we extract microphone channel 0 for both SFT and RLVR.

The evaluation corpus consists of real-world, scripted conversations recorded through the wearable device for live translation scenario. The corpus contains $1{,}181$ five-channel recordings between strictly 2 speakers speaking different languages, across eight language pairs. They are German, Spanish, French, and Italian paired bidirectionally with English. The average duration per conversation is around $210$ seconds.

\vspace{-0.6em}
\subsubsection{Training Setup}
\vspace{-0.2em}
SFT uses 64 H100 GPUs, a global batch size of 64, a maximum sequence length of 16,384, and a peak learning rate of $1\!\times\!10^{-5}$ with cosine decay and 20 warm-up steps. The SFT runs for 450 steps. RLVR initializes from the last checkpoint of the SFT job and also uses it as the frozen reverse-KL reference. RLVR uses 120 H100 GPUs, 512 rollouts per batch, a maximum sequence length of 4,096, the same learning rate, and up to $1{,}000$. Each prompt produces eight samples at $T=1.0$ and top-$p=1.0$. Group-normalized rewards optimize group-normalized policy-gradient objective with full-distribution reverse KL of weight $0.1$; non-finite, effectively zero-variance, and truncated groups are removed. We set $\alpha_{\mathrm{SAE}}=\alpha_{\mathrm{DR}}=0.2$. Unless varied below, delay-controlled runs use a $700\,\text{ms}$ onset buffer, and a $400\,\text{ms}$ decay scale.

\vspace{-0.6em}
\subsubsection{Evaluation Setup}
\vspace{-0.2em}
All headline results use whole-conversation streaming recognition on the $1{,}181$ recordings described above. The first channel is extracted for decoding, matching the training. Greedy decoding uses temperature $0$. For the stochastic SFT study, we draw eight samples per recording using temperature $1.0$, top-$p=1.0$, and top-$k=50$. Oracle@8 selects the minimum-SA-WER candidate for each recording and uses the same candidate for latency; it therefore measures latent model capability rather than deployable single-decode performance.

We report fixed-role SA-WER on untouched hypotheses using the official multi-talker scorer, together with wearer and partner WER where relevant. For latency, hypothesis and reference turns are paired through matching blocks of their fixed-role tag sequences. Onset delay is the speaker-tag emission time minus the oracle turn start time. We report pooled $p50$, $p90$, and $p95$ over paired turns, the fraction with negative onset delay, and selected long-tail rates. Negative onset denotes premature emission rather than low latency. Missing, inserted, or misattributed turns remain unpaired, so latency percentiles are interpreted together with SA-WER and turn coverage.

\subsection{Experiment Results}

\subsubsection{From SFT to SA-WER-RL}
Table~\ref{tab:sft_rl} decomposes the SFT and RL models' performance into the error terms targeted by Eq.~\eqref{eq:combined_error}. SA-WER RL reduces overall error from $20.51\%$ to $13.14\%$. More importantly, wearer/partner SAE falls from $3.2\%/5.0\%$ to $0.5\%/0.3\%$, showing that RL suppresses fixed-role inversions. Wearer/partner deletion also decreases from $12.2\%/14.4\%$ to $9.5\%/8.1\%$, which validates explicitly upweighting deletion errors.

\begin{table}[t]
\centering
\caption{Quality decomposition. $\mathrm{SAE}_{r}$ and $D_r$ are normalized by reference words for role $r\in\{0,1\}$; Eq.~\eqref{eq:combined_error} uses their reference-weighted SAE aggregate and $D$. Speaker-level breakdowns were not retained for oracle@8.}
\label{tab:sft_rl}
\vspace{2mm}
\normalsize
\renewcommand{\arraystretch}{1.1}
\setlength{\tabcolsep}{2pt}
\begin{tabular}{lrrrrr}
\hline
System & SA-WER & $\mathrm{SAE}_0$ & $\mathrm{SAE}_1$ & $D_0$ & $D_1$ \\
 & (\%) & (\%) & (\%) & (\%) & (\%) \\
\hline
SFT & 20.51 & 3.2 & 5.0 & 12.2 & 14.4 \\
SFT oracle@8 & 14.61 & -- & -- & -- & -- \\
SA-WER RL & \textbf{13.14} & \textbf{0.5} & \textbf{0.3} & \textbf{9.5} & \textbf{8.1} \\
\hline
\end{tabular}

\vspace{1.5mm}
\setlength{\tabcolsep}{6pt}
\begin{tabular}{lrr}
\hline
SFT sampling statistic & Pass@1 (\%) & Pass@8 (\%) \\
\hline
Exact ($0\%$ SA-WER) & 0.00 & 0.00 \\
$\leq 5\%$ SA-WER & 9.70 & 25.62 \\
$\leq 10\%$ SA-WER & 33.55 & 58.09 \\
\hline
\end{tabular}
\end{table}

Oracle@8 and pass@8 are complementary: oracle@8 reports corpus SA-WER after selecting the lowest-SA-WER candidate for each recording, whereas pass@8 is the probability that at least one of eight candidates meets a specified threshold. Sampling at $T=1$ does not improve the expected SFT output: candidate-corpus SA-WER rises to $24.96\%$. Nevertheless, oracle@8 reaches $14.61\%$, and Table~\ref{tab:sft_rl} shows substantial pass@8 gains at the $5\%$ and $10\%$ thresholds. Thus SFT already assigns probability mass to better transcripts, but individual samples are worse on average and oracle selection is unavailable at inference. SA-WER RL converts this latent capability into greedy behavior, improving over greedy SFT by $7.37$ absolute points ($35.9\%$ relative) and surpassing SFT oracle@8 by $1.47$ points with one decode. However, when we measure the onset delay, its onset $p50$ remains close to SFT ($0.918$ versus $0.906\,\text{s}$), confirming that this gain is only for the fixed-role quality objective and that latency requires an explicit reward.

\subsubsection{Effect of Onset-Delay Regularization}
\begin{figure}[t]
\centering
\includegraphics[width=\columnwidth]{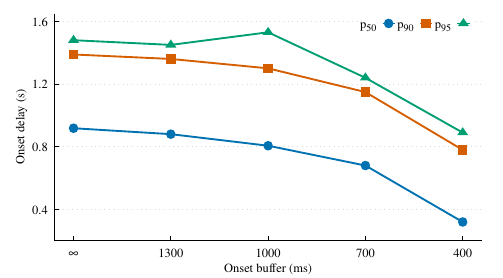}
\caption{Pooled onset-delay percentiles for the same checkpoints. Smaller buffers monotonically improve $p50$ and $p90$; $p95$ is less stable.}
\label{fig:onset_buffer_trend}
\end{figure}

Table~\ref{tab:delay_quality} and Fig.~\ref{fig:onset_buffer_trend} separate recognition quality from latency for representative checkpoints. By Eq.~\eqref{eq:delay_score}, onset credit is constant for $d\leq b$ and decays exponentially beyond buffer $b$ with $\tau=400\,\text{ms}$. The no-delay baseline has $\lambda=0$, so $M_{\mathrm{delay}}=1$ regardless of emission time. Under its observed $p50=0.918\,\text{s}$, the onset credit would be $1.00/0.27$ for buffers of $1{,}300/400\,\text{ms}$. Smaller buffers therefore expose more of the baseline delay distribution to reward decay and provide stronger pressure for earlier emission. Quality remains near the SA-WER-only baseline across the stable $\lambda=0.2$ runs. The $400\,\text{ms}$ model obtains the lowest SA-WER and self/other deletion rates, while the diagnostic $1{,}000\,\text{ms}$ run is degraded mainly by partner attribution error.

\begin{table}[t]
\centering
\caption{Recognition-quality decomposition for the onset-buffer sweep ($\lambda=0.2$ except the no-delay baseline). The $1{,}000$-ms run is diagnostic.}
\label{tab:delay_quality}
\vspace{1.5mm}
\normalsize
\renewcommand{\arraystretch}{1.05}
\setlength{\tabcolsep}{2pt}
\begin{tabular}{lrrrrr}
\hline
Buffer & SA-WER & $\mathrm{SAE}_0$ & $\mathrm{SAE}_1$ & $D_0$ & $D_1$ \\
(ms) & (\%) & (\%) & (\%) & (\%) & (\%) \\
\hline
None & 13.14 & 0.5 & \textbf{0.3} & 9.5 & 8.1 \\
1,300 & 13.41 & \textbf{0.3} & 1.1 & 9.6 & 8.2 \\
1,000 & 15.51 & 1.1 & 4.8 & 9.7 & 8.1 \\
700 & 13.36 & 0.5 & 1.3 & 9.2 & 7.8 \\
400 & \textbf{12.70} & \textbf{0.3} & 1.0 & \textbf{8.8} & \textbf{7.4} \\
\hline
\end{tabular}
\end{table}

Figure~\ref{fig:onset_buffer_trend} shows that reducing the buffer from $1{,}300$ to $400\,\text{ms}$ lowers pooled $p50$ from $0.880$ to $0.320\,\text{s}$ and $p90$ from $1.360$ to $0.780\,\text{s}$. Relative to SA-WER-only RL, the $400\,\text{ms}$ model reduces $p50/p90/p95$ by $65\%/44\%/40\%$, but raises negative onset from $2.11\%$ to $5.49\%$. Its pooled $p95$ is $0.890\,\text{s}$. Thus, smaller buffers reliably accelerate typical turns with stronger regularization, while premature emission remains checkpoint-selection safeguards.

\subsubsection{Delay-Reward Weight Ablation}
Table~\ref{tab:delay_weight} isolates reward strength at a fixed $700\,\text{ms}$ onset buffer. Increasing $\lambda$ from $0.05$ to $0.2$ lowers onset $p50$ by $170\,\text{ms}$ and both $p90$ and $p95$ by approximately $190\,\text{ms}$. SA-WER changes by only $0.19$ absolute points, while the negative-onset rate remains nearly unchanged. The stronger bounded regularizer therefore supplies a materially clearer latency signal without a comparable degradation in recognition quality. Together with the buffer sweep, this motivates $\lambda=0.2$ for the lowest-latency operating point while preserving SA-WER as the dominant objective through the multiplier's $16.7\%$ maximum reward reduction.

\begin{table}[t]
\centering
\caption{Delay-reward weight ablation at a fixed $700\,\text{ms}$ onset buffer.}
\label{tab:delay_weight}
\vspace{2mm}
\normalsize
\renewcommand{\arraystretch}{1.1}
\setlength{\tabcolsep}{3.5pt}
\begin{tabular}{lrrrrr}
\hline
$\lambda$ & SA-WER & $p50$ & $p90$ & $p95$ & Neg. \\
 & (\%) & (s) & (s) & (s) & (\%) \\
\hline
0.05 & \textbf{13.17} & 0.850 & 1.336 & 1.427 & \textbf{2.46} \\
0.20 & 13.36 & \textbf{0.680} & \textbf{1.149} & \textbf{1.240} & 2.50 \\
\hline
\end{tabular}
\end{table}

\vspace{-10pt}
\section{Conclusion}
\label{sec:conclusion}
We presented an RLVR approach for low-latency LLM-based streaming multi-talker ASR. A fixed-role SA-WER reward directly optimizes recognition and speaker attribution, while a bounded, coverage-aware onset-delay regularizer encourages timely turn emission without allowing latency to dominate quality. Our results show that sequence-level verifiable rewards can jointly improve fixed-role recognition quality and streaming responsiveness without modifying the underlying model architecture. Future work will extend evaluation to unscripted conversations and broader language coverage while improving robustness to premature emissions and direction-specific latency tails.



\bibliographystyle{IEEEbib}
\bibliography{strings,refs}

\end{document}